\documentclass[10pt,twocolumn,letterpaper]{article}

\pdfoutput=1

\usepackage{iccv}
\usepackage{times}
\usepackage{epsfig}
\usepackage{graphicx}
\usepackage{amsmath}
\usepackage{amssymb}
\usepackage{makecell}       

\usepackage[numbers,sort]{natbib}
\usepackage{multirow}
\usepackage{fixltx2e}
\usepackage{url}

\usepackage[pagebackref=true,breaklinks=true,letterpaper=true,colorlinks,bookmarks=false]{hyperref}

\iccvfinalcopy 

\def\iccvPaperID{555} 

\begin{document}

\newcommand{\br}{\mbox{$\bf r$}}			
\newcommand{\tc}{\mbox{$\tau$}}				
\newcommand{\sos}{\mbox{$\mathcal{S}$}}		
\newcommand{\nsos}{\mbox{$s$}}				

\newcommand{\todo}[1]{{\color{red} #1}}		
\newcommand{\corr}[1]{{\color{blue} #1}}	


\title{Unsupervised Object Discovery and Tracking in Video Collections}

\author{Suha Kwak\textsuperscript{1,}\thanks{WILLOW project-team, D\'epartement d'Informatique de l'Ecole Normale
Sup\'erieure, ENS/Inria/CNRS UMR 8548.} \quad\quad Minsu Cho\textsuperscript{1,}\footnotemark[1]  \quad\quad Ivan Laptev\textsuperscript{1,}\footnotemark[1] \quad\quad Jean Ponce\textsuperscript{2,}\footnotemark[1] \quad\quad Cordelia Schmid\textsuperscript{1,}\thanks{LEAR project-team, Inria Grenoble Rh\^one-Alpes, France.} \vspace*{0.2cm}\\
{\textsuperscript{1}Inria \quad\quad\quad \textsuperscript{2}\'{E}cole Normale Sup\'erieure / PSL Research University} \vspace*{0.2cm}\\
}

\maketitle


\begin{abstract}
This paper addresses the problem of automatically localizing dominant objects as spatio-temporal tubes in a noisy collection of videos with minimal or even no supervision. 
We formulate the problem as a combination of two complementary processes: discovery and tracking. The first one establishes correspondences between prominent regions across videos, and the second one associates successive similar object regions within the same video.  
Interestingly, our algorithm also discovers the implicit topology of frames associated with instances of the same object class across different videos, a role normally left to supervisory information in the form of class labels
in conventional image and video understanding methods. Indeed, as
demonstrated by our experiments, our method can handle video
collections featuring multiple object classes, and substantially outperforms the state of the art in colocalization, 
even though it tackles a broader problem with much less supervision.
\end{abstract}


\section{Introduction}
 
Visual learning and interpretation is traditionally formulated as a supervised classification problem, with manually selected bounding boxes acting as (strong) supervisory signal~\cite{imagenet_cvpr09,pascal-voc-2007}. 
To reduce human effort and subjective biases in manual annotation, recent work has addressed the discovery and localization of objects from weakly-annotated or even unlabelled datasets~\cite{Cinbis2014,shi2013bayesian,cho2015,Deselaers:2010he,Tang14}. 
However, this task is difficult and most approaches today still lay significantly behind strongly-supervised methods.
With the ever growing popularity of video sharing sites such as YouTube, 
recent research has started to handle the similar task in videos~\cite{Sharir:2012if,Wang2014,Prest2012,Joulin14}, and has shown that exploiting the space-time structure of the world, which is absent in static images, \eg, motion information, may be crucial for achieving object discovery or localization with less supervision.

\begin{figure*}[t]
\center
\includegraphics[width=0.9\linewidth]{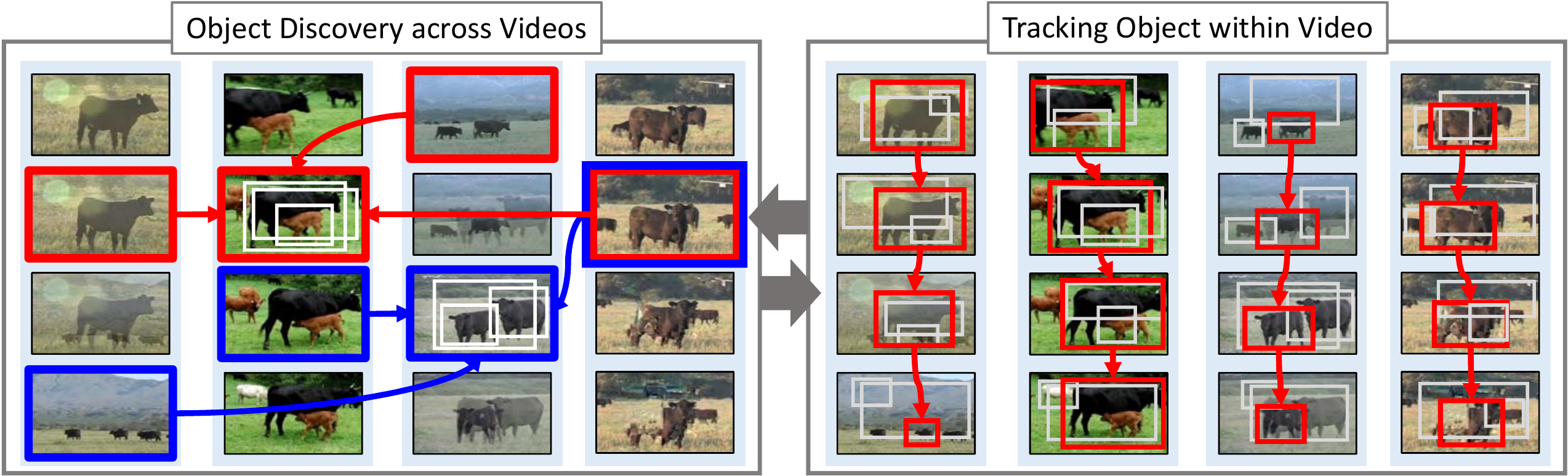}
\caption{
Given a noisy collection of videos, dominant objects are automatically localized as spatio-temporal tubes.
The discovery process establishes correspondences between prominent regions across videos (left), and the tracking process associates similar object regions within the same video (right).  
(Best viewed in color.)}
\label{fig:intro}
\end{figure*}


This paper addresses the problem of spatio-temporal object localization in videos with minimal supervision or even no supervision. Given a noisy collection of videos with multiple object classes, dominant objects are identified as spatio-temporal tubes for each video (Fig.~\ref{fig:intro}).
We formulate the problem as a combination of two complementary processes: object {\em discovery} and {\em tracking}. 
In our daily experience, salient motion often primes us to recall similar visual patterns as an object from our memory, and such recalled patterns help us to localize the object over time.
Likewise, object discovery, whose aim is to establish correspondences between regions depicting similar objects in frames of different videos, is closely connected to object tracking, whose aim is to associate target objects in consecutive video frames. 
Building upon recent advances in efficient matching~\cite{cho2015} and tracking~\cite{Pirsiavash2011}, 
we combine region matching across different videos and region tracking within each video into a joint optimization framework. 
We demonstrate that the proposed method substantially outperforms the state of the art in colocalization~\cite{Joulin14} on the YouTube-Object dataset, even though it tackles a broader problem with much less supervision.



\subsection{Related work}

Our approach combines object discovery and tracking. 
The discovery part establishes correspondences between frames across videos to detect object candidates.
Similar approaches have been proposed for salient region detection~\cite{kim2009unsupervised}, image cosegmentation~\cite{Wang2014unsupervised,Wang2013functional}, and image colocalization~\cite{cho2015}.
Conventional object tracking methods~\cite{Yilmaz2006} usually require annotations for at least one frame~\cite{Wu2013,Hare2011,Hong2013}, or object detectors trained for target classes in a supervised manner~\cite{Pirsiavash2011,Benfold2011,Breitenstein2011}.  Our method does not require such supervision and instead alternates discovery and tracking of object candidates.

The problem we address is closely related to video object colocalization~\cite{Joulin14,Prest2012}, whose goal is to localize the common object in a video collection.
Prest~\etal~\cite{Prest2012} generate spatio-temporal tubes of object candidates, and select one of these per video through energy minimization.
Since the candidate tubes rely only on clusters of point tracks~\cite{Brox2010}, this approach is not robust against noisy tracks and incomplete clusters.
Joulin \etal~\cite{Joulin14} extend the image colocalization framework~\cite{Tang14} for videos using an efficient optimization approach.
This method does not explicitly consider correspondences between frames from different videos, which are shown to be essential for robust localization of common objects in our experiments of Section~\ref{sec:exp_coloc}.

Our setting is also related to object segmentation or cosegmentation in videos.
For video object segmentation, clusters of long-term point tracks have been used~\cite{Brox2010,Ochs2011,Ochs2012}, while assuming that points from the same object have similar tracks.
In~\cite{Papazoglou2013,Lee2011key}, appearances of potential object and background are modeled and combined with motion information for the task.
These methods produce results for individual videos and do not investigate relationships between videos and the objects they contain. 
Video object cosegmentation aims to segment a detailed mask of common object out of videos. 
This problem has been addressed with weak supervision such as object class per video~\cite{Tang2013} and additional labels for a few frames that indicate whether the frames contain target object or not~\cite{Wang2014}.


\subsection{Proposed approach}
\label{sec:proposed_approach}

We consider a set of videos $v$, each consisting of $T$ frames
(images) $v_t$ ($t=1,\ldots,T$), and denote by $R(v_t)$ a set of
candidate regions identified in $v_t$ by some separate bottom-up
proposal process~\cite{Manen2013}. 
Every region proposal is represented by a box in this paper.
We also associate with $v_t$ a {\em matching
neighborhood} $N(v_t)$ formed by the $k$ closest frames $w_u$ among
all videos $w\neq v$, according to a robust criterion based on
probabilistic Hough matching (see~\cite{cho2015} and Section~\ref{sec:appearance_saliency}).
The network structure defined by $N$ links {\em frames} across {\em
different} videos (Fig.~\ref{fig:intro}, left). We also link {\em regions} in successive frames of
the {\em same} video, so that $r_t$ in $R(v_t)$ and $r_{t+1}$ in
$R(v_{t+1})$ are {\em tracking neighbors} when there exists some point
track originating in $r_t$ and terminating in $r_{t+1}$ (Fig.~\ref{fig:intro}, right).
A {\em spatio-temporal tube} is any sequence $r=[r_1,\ldots,r_T]$ of
temporal neighbors in the same video.  Our goal is to find, for every
video $v$ in the input collection, the top tube $r$
according to the criterion
\begin{equation}
\Omega_v(r)=
\sum_{t=1}^T \varphi[r_t,v_t,N(v_t)] +\lambda
\sum_{t=1}^{T-1} \psi (r_t,r_{t+1}),\label{eq:objective}
\end{equation}
where $\varphi[r_t,v_t,N(v_t)]$ is a measure of confidence for $r_t$
being an object (foreground) region, given $v_t$ and its matching
neighbors, and $\psi(r_t,r_{t+1})$ is a measure of temporal
consistency between $r_t$ and $r_{t+1}$. 

As will be shown in the sequel, given the matching network structure
$N$, finding the top tube (or for that matter the top $p$ tubes) for
each video can be done efficiently using dynamic programming.  Note
that both the matching and tracking network structures are a priori
fixed. However, the matching network is huge, every frame in a video
being a priori linked to all other frames in all other videos, and, as
will be shown in Section \ref{sec:appearance_saliency}, computing the matching score between two
frames is itself nontrivial. We therefore choose instead to use an
iterative process, alternating between steps where $N$ is fixed and
the top $k$ tubes are computed for each video, with steps where the
top $k$ tubes are fixed, and used to update the matching
network. After a few iterations, we stop, and finally pick the top
scoring tube for each video. We dub this iterative process a {\em
discovery and tracking} procedure since finding the tubes maximizing
foreground confidence across videos is akin to unsupervised
object discovery~\cite{cho2015,ICCV/SivicREZF05,Russell06,grauman2006unsupervised,faktor2012clustering}, 
whereas finding the tubes maximizing temporal
consistency within a video is similar to object tracking~\cite{Yilmaz2006,Wu2013,Hare2011,Pirsiavash2011,Benfold2011,Breitenstein2011}.

Interestingly, because we update the matching neighborhood structure
at every iteration, our discovery and tracking procedure does much more
than finding the spatio-temporal tubes associated with dominant
objects: It also discovers the implicit neighborhood structure of
frames associated with instances of the same class, which is a role
normally left to supervisory information in the form of class labels
in conventional image and video understanding methods. Indeed, as
demonstrated by our experiments, our method can handle video
collections featuring multiple object classes with minimal or zero
supervision (it is, however, limited for the time being to one object
instance per frame).

We describe in the next two sections our foreground confidence and temporal consistency terms of Eq.~\eqref{eq:objective}, before describing in Section \ref{sec:algorithm_detail} our discovery and tracking algorithm, presenting experiments in Section \ref{sec:exp}, and concluding in Section \ref{sec:conclusion} with brief remarks about future work.

\section{Foreground confidence}

Our foreground confidence term is defined as a weighted sum of appearance- and motion-based confidences:
\begin{equation}
\varphi[r_t,v_t,N(v_t)] = \varphi_{\rm a}[r_t,v_t,N(v_t)] + \alpha \ \varphi_{\rm m}(r_t).
\end{equation}
For the appearance-based term denoted by $\varphi_{\rm a}$, we follow~\cite{cho2015} and use a {\it standout score} based on region matching confidence. For the motion-based term denoted by $\varphi_{\rm m}$, we build on long-term point track clusters~\cite{Brox2010} and propose a {\it motion coherence score} that measures how well the box region aligns with motion clusters. 

\subsection{Appearance-based confidence}
\label{sec:appearance_saliency}

Foreground object regions are likely to match each other across videos with similar objects, and a region tightly bounding a foreground object stands out over the background.
Recent work on unsupervised object discovery in image collections~\cite{cho2015} implements this concept through a {\it standout score} based on a region matching algorithm, called probabilistic Hough matching (PHM). Here we extend the idea to video frames.
 

PHM is an efficient region matching algorithm which generates scores for region matches using appearance and geometric consistency. 
Assume two sets of region proposals have been extracted from $v_t$ and $v_u$: $R_t = R(v_t)$ and $R_u=R(v_u)$.
Let $r_t=(f_t, l_t) \in R_t$ be a region with its $8 \times 8$ HOG descriptor $f_t$~\cite{hariharan2012discriminative,dalal2005histograms} and its location $l_t$, \ie, position and scale.
The score for match $m=(r_t,r_u)$ is decomposed into an appearance term $m_{\rm a}=(f_t, f_u)$ and a geometry term $m_{\rm g}=(l_t, l_u)$.
Let $x$ denote the location offset of a potential object common to $v_t$ and $v_u$.
Given $R_t$ and $R_u$, PHM evaluates the match score $c(m| R_t,R_u)$ by combining the Hough space vote $h(x | R_t, R_u)$ and the appearance similarity in a pseudo-probabilistic way:      
\begin{eqnarray}
c ( m | R_t, R_u ) &=& p(m_{\rm a}) \sum_x p(m_{\rm g} | x) h(x | R_t, R_u), \label{eq:match_conf}\\
h( x | R_t, R_u)  &=& \sum_{m} p(  m_{\rm a} ) p(m_{\rm g} | x ), \label{eq:hough_conf}
\end{eqnarray}
where $p(m_{\rm a})$ is the appearance-based similarity between two descriptors $f_t$ and $f_u$, and 
$p(m_{\rm g} | x)$ is the likelihood of displacement $l_t-l_u$, which is defined as a Gaussian distribution centered on $x$. As noted in~\cite{cho2015}, this can be seen as a combination of bottom-up Hough space voting (Eq.~\eqref{eq:hough_conf}) and top-down confidence evaluation (Eq.~\eqref{eq:match_conf}). 
Given neighbor frames $N(v_t)$ where an object in $v_t$ may appear, 
the region saliency is defined as the sum of max-pooled match scores from $R'_u$ to $r$:
\begin{equation}
g(r_t | R_t, R_u) = \sum_{v_u \in N(v_t)} \max_{r_u \in R_u} \ c \big( (r_t, r_u) | R_t, R_u \big). \label{eq:region_conf_ext}
\end{equation}
We omit the given terms $R_t$ and $R_u$ in function $g$ for brief notation afterwards. 
The region saliency $g(r_t)$ is high when $r$ matches the neighbor frames well in terms of both appearance and geometric consistency. 
While useful as an evidence for foreground regions, the region saliency of Eq.~\eqref{eq:region_conf_ext} may be higher on a part than a whole object because part regions often match more consistently than entire object regions. To counteract this effect, a standout score measures how much the region $r_t$ ``stands out'' from its potential backgrounds in terms of region saliency: 
\begin{eqnarray}
&&\!\!\!\!\!  s(r_t) = g(r_t) - \max_{r_\text{b} \in B(r_t)} g(r_\text{b}), \nonumber \\
&&\!\!\!\!\!\!\!\!\!\!\!\!\!\!\!   \textrm{s.t.} \ \ \  B(r_t) = \{ r_\text{b} | r_t \subsetneq r_\text{B}, r_\text{b} \in R_t \},
\end{eqnarray}
where $r_t \subsetneq r_\text{b}$ indicates that region $r_t$ is contained in region $r_\text{b}$. 
As can be seen from Eq.~\eqref{eq:region_conf_ext}, the standout score $s(r_t)$ evaluates a foreground likelihood of $r_t$ based on region matching between frame $v_t$ and its neighbor frames $N(v_t)$. Now we denote it more explicitly using $s\bigl( r_t | v_t, N(v_t) \bigr)$.
The appearance-based foreground confidence for region $r_t$ is defined as the standout score of $r_t$:
\begin{eqnarray}
\varphi_{\rm a} [r_t,v_t,N(v_t)] = s \bigl( r_t | v_t, N(v_t) \bigr). 
\label{eq:app_region_conf}
\end{eqnarray}
In practice, we rescale standout scores to cover $[0, 1]$ at each frame. 

\subsection{Motion-based confidence}
\begin{figure}[t]
\centering
\begin{minipage}{1.0\linewidth}
\centering
\includegraphics[width=0.99\linewidth]{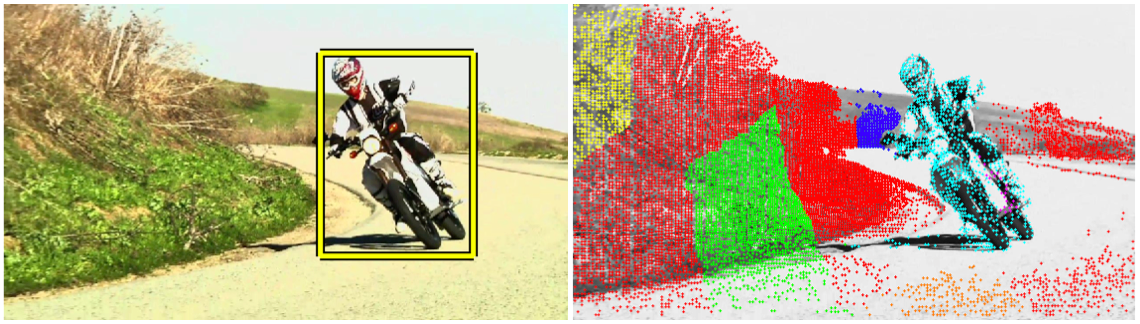}
\end{minipage}\vskip0.05cm
\begin{minipage}{1.0\linewidth}
\centering
\small{(a) Video frame and its color-coded motion clusters.}
\end{minipage}\vskip0.1cm
\begin{minipage}{1.0\linewidth}
\includegraphics[width=0.99\linewidth]{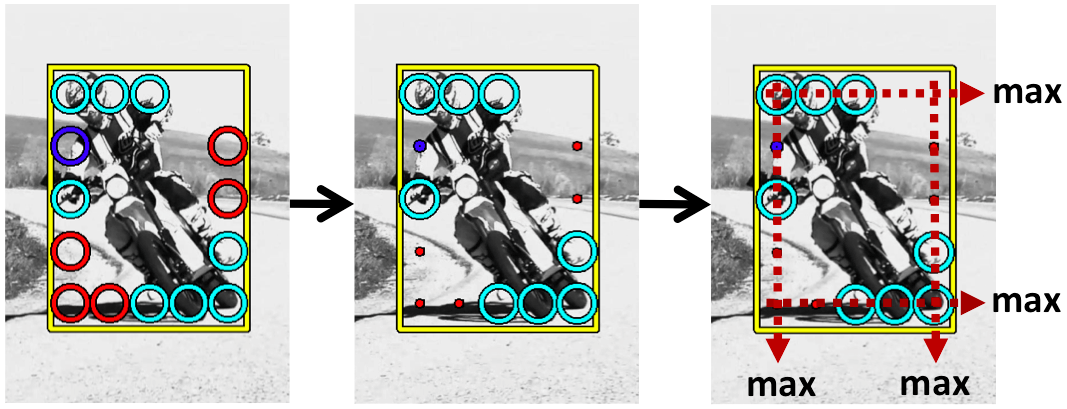}
\end{minipage}\vskip0.05cm
\begin{minipage}{1.0\linewidth}
\centering
\small{(b) Measuring the motion coherence score for a box region.}
\end{minipage}\vskip0.1cm
\begin{minipage}{1.0\linewidth}
\includegraphics[width=0.99\linewidth]{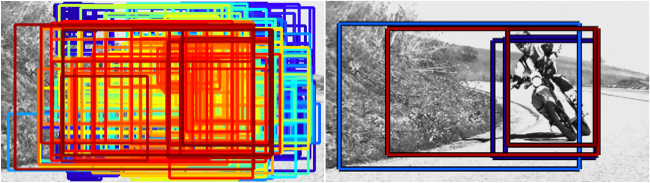}
\end{minipage}\vskip0.05cm
\begin{minipage}{1.0\linewidth}
\centering
\small{(c) Heat map of the scores and the top 5 boxes.}
\end{minipage}\vskip0.1cm
\caption{ Motion-based region confidence. (a) Given a video clip, its motion clusters are computed for each frame~\cite{Brox2010}. The example shows a frame (left) and its motion cluster with color coding (right). (b) Given a box region (yellow), the motion coherence score  for the box is computed in three steps: box-boundary binning (left), cluster weighting (middle), and edge-wise max pooling (right). For the details, see text. (c) Heat map of the motion coherence scores (left) and the top 5 box regions with the best scores (right). (Best viewed in color.) }
\label{fig:region_saliency_mot}
\end{figure}

Motion is an important cue for localizing moving objects in videos and differentiating them from the background~\cite{Papazoglou2013}. 
To exploit this information, we propose the {\em motion coherence score} as another foreground confidence measure, which is built on clusters of long-term point tracks~\cite{Brox2010}.
Since the motion clusters incorporate long-term spatio-temporal coherence, they are more ``global'' than conventional optical flows and long-term tracks.
Using the motion clusters, we propose to compute the motion coherence score for a box region in three steps: (1) edge motion binning, (2) motion cluster weighting, (3) edge-wise max pooling. 
First, we divide a box region into $5 \times 5$ cells, and construct bins along its edges as illustrated in Fig.~\ref{fig:region_saliency_mot}. Then, for each bin $b$, we assign its cluster label $l_b$ by majority voting using the tracks that fall into the bin. 
Second, we compute a weight for each motion cluster:
\begin{equation}
w(l) =  \frac{\# \text{ of tracks of cluster $l$ within the box} }{\# \text{ of all tracks of cluster $l$ in the frame} }, 
\end{equation}
evaluating how much of the motion cluster the box includes, compared to the entire frame. The weight is assigned to the corresponding bin, and suppresses the effect of background clusters in the bins. Third, we select the bin with the maximum cluster weight along each edge, and define the sum of the weights as the motion coherence score for the box:
\begin{equation}
\varphi_{\rm m}(r_t) =  \sum_{e \in \{\rm L, R, T, B \}}{ \max_{b\in E_e}{w(l_b)}},
\end{equation}
where $e$ represents one of four edges of box region (left, right, top, bottom), $E_e$ a set of bins on the edge, and $l_b$ the cluster label of bin $b$. 
This score is designed to be high for a box region that contacts with motion cluster boundaries (edge-wise max pooling) and contains the entire clusters (motion cluster weighting). 
Note that in most cases an object does not fill the entire area of its bounding box, but only touches the four edges (\eg, round objects).
On this account, edge-wise max pooling provides a more robust score than average pooling on entire cells. If the box does not touch any motion cluster boundary, the score becomes small since some tracks of pooled clusters lay outside of the box. 
This motion coherence score is useful to discover moving objects in video frames, and acts a complementary cue to the standout score in Section~\ref{sec:appearance_saliency}.


\section{Temporal consistency}

Regions with high foreground confidences may turn out to be temporally inconsistent. 
They can be misaligned due to imperfect confidence measures and ambiguous observations.
Also, given multiple object instances of the same category, foreground regions may correspond to different instances in a video. 
Our temporal consistency term is used to handle these issues so that selected spatio-temporal tubes are more stable and consistent temporally.
We exploit both appearance- and motion-based evidences for this purpose. 
We denote by $\psi_{\rm a}(r_t, r_{t+1})$ and $\psi_{\rm m}(r_t, r_{t+1})$ appearance- and motion-based terms, respectively.  
The consistency term of Eq.~\eqref{eq:objective} is obtained as 
\begin{equation}
\psi(r_t, r_{t+1}) = \psi_{\rm a}(r_t, r_{t+1}) + \psi_{\rm m}(r_t, r_{t+1}).
\end{equation}
We describe these terms in the following subsections.

\subsection{Appearance-based consistency}

We use appearance similarity between two consecutive regions as a temporal consistency term.
Region $r_t$ is described by an $8 \times 8$ HOG descriptor $f_t$, as in Section~\ref{sec:appearance_saliency}, and the appearance-based consistency is defined as the opposite of the distance between descriptors: 
\begin{equation}
\psi_{\rm a}(r_t, r_{t+1}) = -|| f_t - f_{t+1} ||_2, 
\end{equation}
which is rescaled in practice to cover $[0, 1]$ at each frame. 

\subsection{ Motion-based consistency}

\begin{figure}[t]
\center
\includegraphics[width=0.99\linewidth]{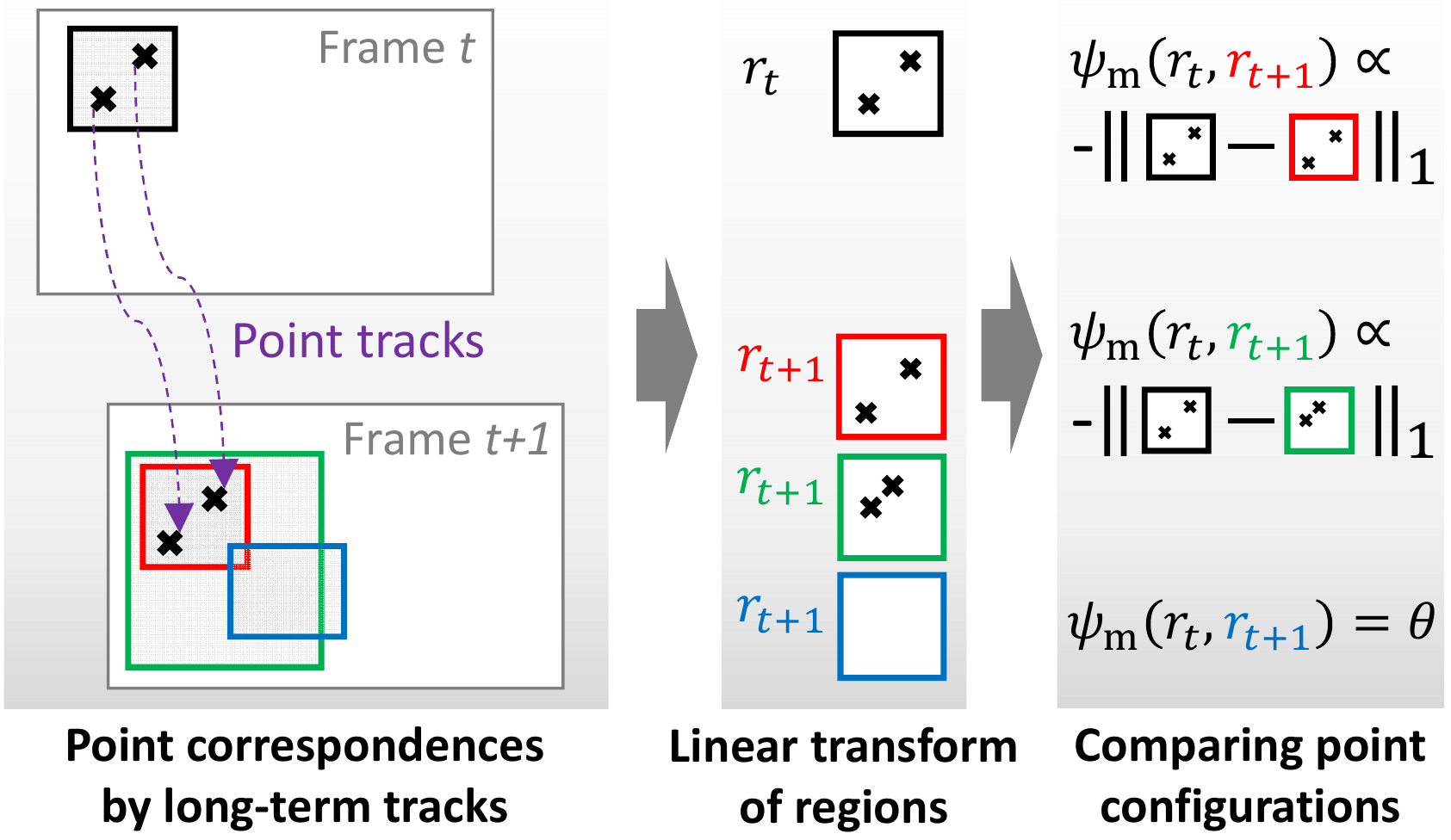}
\caption{ 
Motion-based temporal consistency. We compare two sets of corresponding points in consecutive regions by transforming them into a unit square from the regions. The configuration of points does not align with each other unless two regions match well (\eg, {\it black} and {\it green}).
The motion-based consistency uses the sum of distances between the corresponding points in the transformed domain. If two regions share no point track, we assign a constant $\theta$ as the consistency term. (Best viewed in color.)
}
\label{fig:temp_smooth_mot}
\end{figure}

Two consecutive regions $r_t$ and $r_{t+1}$ associated with the same object typically share the same point tracks, and configurations of the points in the two regions should be similar. Long-term point tracks~\cite{Brox2010} provide correspondences for such points across frames, which we exploit to measure the motion-based consistency between a pair of regions.

To compare the configurations of shared point tracks, we linearly transform each box region and internal point coordinates  
into a unit square with edge length $1$, as illustrated in Fig.~\ref{fig:temp_smooth_mot}.
Using the transformed coordinates, we can compare the point configurations up to affine variation between the regions.
Let $p$ be an individual point track and $p_t$ the coordinate of $p$ at frame $t$. 
Then, the coordinate of $p$ transformed by region $r_t$ is denoted by $\tau(p_t | {r_t})$.
If two consecutive regions $r_t$ and $r_{t+1}$ cover the same object and share a point track $p$, $\tau(p_t|r_t)$ and $\tau(p_{t+1}|r_{t+1})$ should be close to each other.
The motion-based consistency $\psi_{\rm m}(r_t, r_{t+1})$ reflects this observation.
Let $P_{r_t}$ be the set of points occupied by region $r_t$. 
The motion-based consistency is defined as 
\begin{equation}
\psi_{\rm m}(r_t, r_{t+1}) = - \hspace{-0.5cm} \sum_{p \in P_{r_t} \cap P_{r_{t+1}}} \hspace{-0.5cm} \frac{ || \tau(p_t|r_t) - \tau(p_{t+1}|r_{t+1}) ||_1}{2 \mid P_{r_t} \cap P_{r_{t+1}} \mid }. 
\end{equation}
If $r_t$ and $r_{t+1}$ share no point track, we assign a constant value $\psi_{\rm m}(r_t, r_{t+1}) = \theta$, which is smaller than -1, the minimum value of $\psi_{\rm m}(r_t, r_{t+1})$, to penalize transitions between regions having no point correspondence.
This is a bit more inclusive than described in Section~\ref{sec:proposed_approach}, for added robustness.
Through this consistency term, we can measure variations in spatial position, aspect ratio, and scales between regions at the same time.


\section{Discovery and tracking algorithm}
\label{sec:algorithm_detail}

We initialize each tube $r$ as an entire video (a sequence of entire frames), 
and alternate between (1) updating the neighborhood structure across videos and (2) optimizing ${\Omega_{v}(r)}$ within each video. 
The intuition is that better object discovery may lead to more accurate object tracking, and vice versa. 
These two steps are repeated for a few iterations until (near-) convergence. In our experiments, using more than 5 iterations does not improve performance. 
The number of neighbors for each frame is fixed as $k = 10$. 
The final result is obtained by selecting the best tube for each video at the end. As each video is independently processed at each iteration, the algorithm is easily parallelized. 

\vspace{-0.5cm}
\paragraph{Network update.} 
Given a localized tube $r$ fixed for each video, we update the neighborhood structure $N$ by $k$ nearest neighbor retrieval for each localized object region. 
At the first iteration, the nearest neighbor search is based on distances between GIST descriptors~\cite{torralba2008small} of frames as the tube $r$ is initialized as the entire video.
From the second iteration, the metric is defined as the appearance similarity between potential object regions localized at the previous iteration.
Specifically, we select top 20 region proposals inside the potential object regions according to region saliency (Eq.~\eqref{eq:region_conf_ext}), and perform PHM between those small sets of regions.
The similarity is then computed as the sum of all region saliency scores given by the matching.
This selective region matching procedure allows us to perform efficient and effective retrieval for video frames.

\vspace{-0.48cm}
\paragraph{Object relocalization.}
Given the neighborhood structure $N$, we optimize the objective of Eq.~\eqref{eq:objective} for each video $v$. 
To exploit the tubes localized at the previous iteration, we confine region proposals in neighbor frames to those contained in the localized tube of the frames. This is done in Eq.(\ref{eq:app_region_conf}) by substituting the neighbor frames of each frame $v_t$ with the regions $r_u$ localized in the frames: 
set $ w_u = r_u$ for all $w_u$ in $N(v_t)$. 
Before the optimization, we compute foreground confidence scores of region proposals, and select the top 100 among these according to their confidence scores.
Only the selected regions are considered during optimization for efficiency.
The objective of Eq.(\ref{eq:objective}) is then efficiently optimized by dynamic programming (DP)~\cite{Pirsiavash2011}. 
Note that using the $p$ best tubes ($p=5$ in all our experiments)
for each video at each iteration except the last one, instead of
retaining only one candidate at each iteration, increases the
robustness of our approach. 
This agrees with the conclusions
of~\cite{cho2015} in the still image domain, and has also been confirmed
empirically by our experiments.
We obtain $p$ best tubes  by sequential DPs, which iteratively remove the best tube and re-run DP again.\footnote{It has been empirically shown in multi-target tracking that sequential DP performs close to the global optimum with greater efficiency than the optimal algorithm~\cite{Pirsiavash2011}.}

\section{Implementation and results}
\label{sec:exp}

Our method is evaluated on the YouTube-Object dataset \cite{Prest2012}, which consists of videos downloaded from YouTube by querying for 10 object classes from P{\small ASCAL} VOC~\cite{pascal-voc-2007}.
Each video of the dataset comes from a longer video and is segmented by automatic shot boundary detection. 
This dataset is challenging since the videos involve large camera motions, view-point changes, encoding artifacts, editing effects, and incorrect shot boundaries. 
Ground-truth boxes are given for a subset of the videos, and one frame is annotated per video for evaluation.
Following~\cite{Joulin14}, our experiments are conducted on all the annotated videos.

We demonstrate the effectiveness of our method through various experiments. 
First, we evaluate our method in the weakly-supervised {\it colocalization} setting, where all videos contain at least one object of the same category.
Our method is also tested in a fully unsupervised mode, where all videos from all classes of the dataset are mixed; we call this challenging setting {\it unsupervised object discovery}.

\subsection{Implementation details}

\paragraph{Key frame selection.}
We sample key frames from each video uniformly with stride 20, and our method is used only on the key frames.
This is because temporally adjacent frames typically have redundant information, and it is time-consuming to process all the frames.
Note that long-term point tracks enable us to utilize continuous motion information although our method works on temporally sparse key frames.
To obtain temporally dense localization results, object regions in non-key frames are estimated by interpolating localized regions in temporally adjacent key-frames.

\vspace{-0.4cm}
\paragraph{Parameter setting.}
The weight for the motion-based confidence $\alpha$ and that for the temporal consistency terms $\lambda$ are set to 0.5 and 2, respectively.
To penalize transitions between regions sharing no point track, $\theta$ is set to -2, smaller than the minimum value of $\psi_{\rm m}$ when two regions share points.
The parameters are fixed for all experiments.

\subsection{Evaluation metrics}

Our method not only discovers and localizes objects, but also reveals the topology between different videos and the objects they contain.
We evaluate our results on those two tasks with different measures.

Localization accuracy is measured using CorLoc~\cite{Prest2012,Joulin14,Papazoglou2013}, which is defined as the percentage of images correctly localized according to the P{\small ASCAL} criterion: $\frac{area( r_{p} \cap r_{gt} )}{area(r_{p} \cup r_{gt} )} > 0.5$, where $r_{p}$ is the predicted region and $r_{gt}$ is the ground-truth.

In the unsupervised object discovery setting, we measure the quality of the topology revealed by our method as well as localization performance.
To this end, we first employ the CorRet metric, originally introduced in~\cite{cho2015}, which is defined in our case as the mean percentage of retrieved nearest neighbor frames that belongs to the same class as the target video.
We also measure the accuracy of nearest neighbor classification, where a query video is classified by the most frequent labels of its neighbor frames retrieved by our method.
The classification accuracy is reported by the top-$k$ error rate, which is the percentage of videos whose ground-truth labels do not belong to the $k$ most frequent labels of their neighbor frames.
All the evaluation metrics are given as percentages.

\begin{table*}[!tbp]
\begin{center}
\caption{CorLoc scores on the YouTube-Object dataset. }
\label{tab:corloc}
\vspace{-0.35cm}
\begin{tabular}{c|cccccccccc|c}
\Xhline{3\arrayrulewidth}
	Method & aeroplane & bird & boat & car & cat & cow & dog & horse & motorbike & train & Avg. \\
\hline
\hline
	Prest \etal~\cite{Prest2012}					& 51.7 & 17.5 & 34.4 & 34.7 & 22.3 & 17.9 & 13.5 & 26.7 & 41.2 & 25.0 & 28.5 \\
	Joulin \etal~\cite{Joulin14}					& 25.1 & 31.2 & 27.8 & 38.5 & 41.2 & 28.4 & 33.9 & 35.6 & 23.1 & 25.0 & 31.0 \\
\hline
	F(A)$^\dagger$									& 38.2 & 67.3 & 30.4 & 75.0 & 28.6 & 65.4 & 38.3 & 46.9 & 52.0 & 25.9 & 46.8 \\
	F(A)+T(M)										& 44.4 & 68.3 & 31.2 & 76.8 & 30.8 & 70.9 & 56.0 & 55.5 & 58.0 & 27.6 & 51.9 \\
	F(A)+T(A,M)										& 52.9 & 72.1 & 55.8 & 79.5 & 30.1 & 67.7 & 56.0 & 57.0 & 57.0 & 25.0 & 55.3 \\
	Ours, full$^\ddagger$							& 56.5 & 66.4 & 58.0 & 76.8 & 39.9 & 69.3 & 50.4 & 56.3 & 53.0 & 31.0 & 55.7 \\
\hline
\hline
	Brox and Malik~\cite{Brox2010}					& 53.9 & 19.6 & 38.2 & 37.8 & 32.2 & 21.8 & 27.0 & 34.7 & 45.4 & 37.5 & 34.8 \\
	Papazoglou and Ferrari~\cite{Papazoglou2013}	& 65.4 & 67.3 & 38.9 & 65.2 & 46.3 & 40.2 & 65.3 & 48.4 & 39.0 & 25.0 & 50.1 \\
\hline
	Ours, full---unsupervised						& 55.2 & 58.7 & 53.6 & 72.3 & 33.1 & 58.3 & 52.5 & 50.8 & 45.0 & 19.8 & 49.9 \\
\Xhline{3\arrayrulewidth}
\multicolumn{10}{l}{\footnotesize{$^\dagger$Our re-implementation of PHM~\cite{cho2015}. \ \ \ \ \ $^\ddagger$Our full method corresponds to F(A,M)+T(A,M).}}
\end{tabular}
\end{center}
\end{table*}

\begin{figure}[!t]
\centering
    \includegraphics[trim = 66mm 109mm 71mm 113mm, clip, width = 0.48 \linewidth]{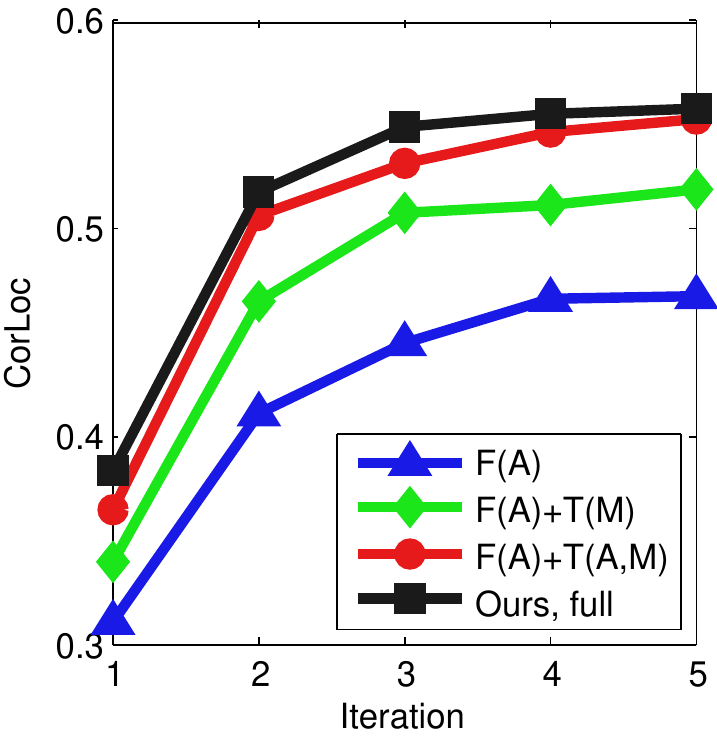} \ \
    \includegraphics[trim = 66mm 109mm 71mm 113mm, clip, width = 0.48 \linewidth]{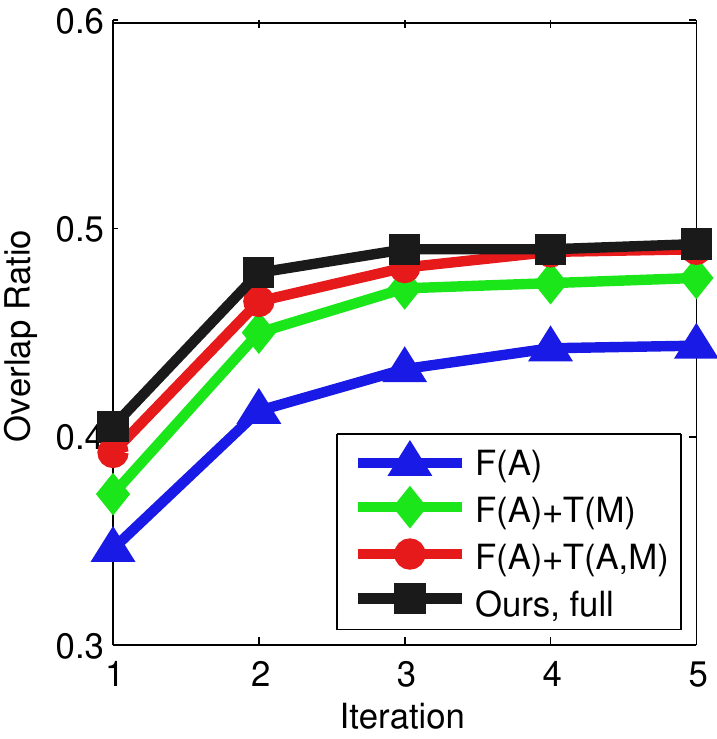} \\
\caption{Average CorLoc scores ({\it left}) and average overlap ratios ({\it right}) versus iterations on the YouTube-Object dataset in the colocalization setting.}
\label{fig:perf}
\end{figure}

\subsection{Object colocalization per class}
\label{sec:exp_coloc}

We compare our method with two colocalization methods for videos~\cite{Prest2012,Joulin14}.
We also compare our method with several of its variants to highlight benefits of each of its components.
Specifically, the components of our method are denoted by combinations of four characters: `F' for foreground confidence, `T' for temporal consistency, `A' for appearance, and `M' for motion.
For example, F(A) means foreground saliency based only on appearance (\ie, $\varphi_a$), and T(A,M) indicates temporal smoothness based on both of appearance and motion (\ie, $\psi_a + \psi_m = \psi$).
Our full model corresponds to F(A,M)+T(A,M).

Quantitative results are summarized in Table~\ref{tab:corloc}.
Our method outperforms the previous state of the art in~\cite{Joulin14} on the same dataset, with a substantial margin. 
Comparing our full method to its simpler versions, we observe that performance improves by adding each of the temporal consistency terms.
The motion-based confidence can damage performance when motion clusters include only a part of object (\eg, ``bird'', ``dog'') and/or background has distinctive clusters due to complex 3D structures (\eg, car, motorbike). However, it enhances localization when the object is highly non-rigid (\eg, ``cat'') and/or is clearly separated from the background by motion (\eg, ``aeroplane'', ``boat''). In the ``train'' class case, where our method without motion-based confidence often localize only a part of long trains,  the motion-based confidence significantly improves localization accuracy. 
Fig.~\ref{fig:perf} illustrates the performance of our method over iterations.
Our full method performs better than its variants at every iteration, and most quickly improves both of CorLoc score and overlap ratio in early stages.

\begin{figure*}[!htbp]
\centering
    \includegraphics[width = 1 \textwidth]{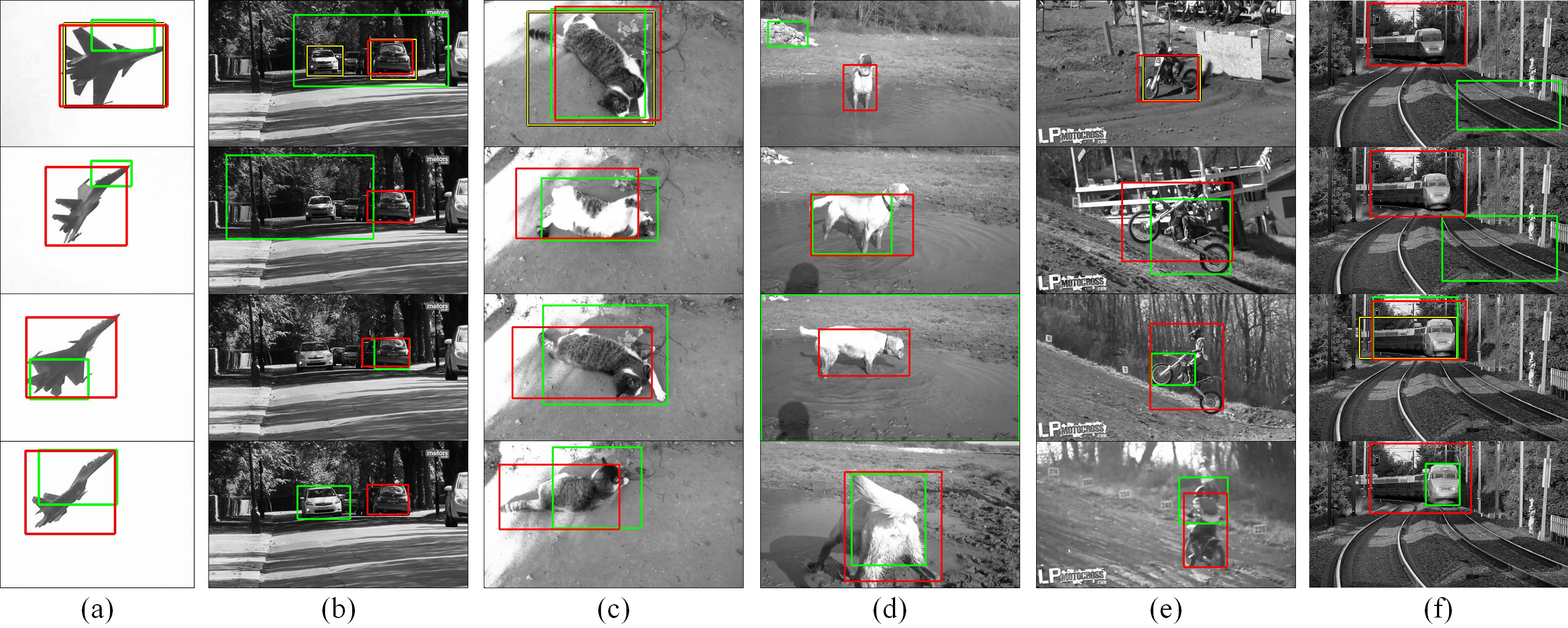}
\caption{Examples of object correctly localized by our full method: ({\it red}) our full method, ({\it green}) our method without motion information, ({\it yellow}) ground-truth localization.
The sequences come from (a) ``aeroplane'', (b) ``car'', (c) ``cat'', (d) ``dog'', (e) ``motorbike'', and (f) ``train'' classes. 
Frames are ordered by time from top to bottom.
The localization results of our full method are spatio-temporally consistent.
On the other hand, the simpler version often fails due to pose variations (a, c--e) or produces inconsistent tracks when multiple target objects exist (b).
More results are included in the supplementary material.
(Best viewed in color.)}
\label{fig:qualitative_coloc}
\end{figure*}

\begin{figure*}[!htbp]
\centering
    \includegraphics[width = 1 \textwidth]{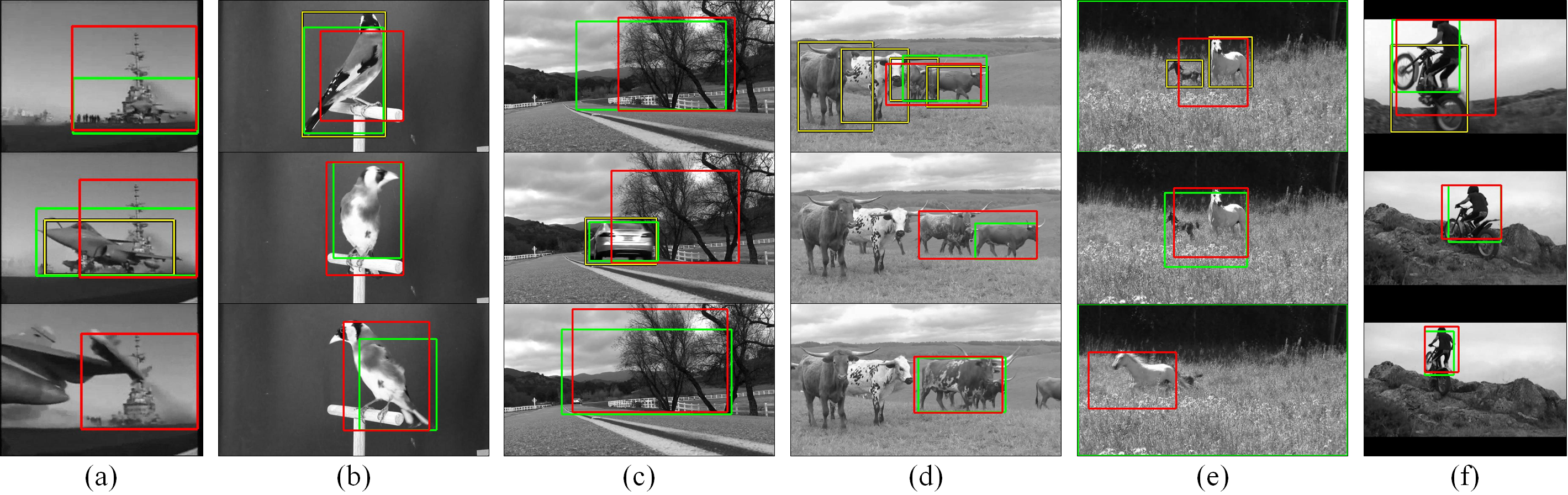}
\caption{Examples incorrectly localized by our full method: ({\it red}) our full method, ({\it green}) our method without motion information, ({\it yellow}) ground-truth localization.
The sequences come from (a) ``aeroplane'', (b) ``bird'', (c) ``car'', (d) ``cow'', (e) ``horse'', and (f) ``motorbike''. 
Frames are ordered by time from top to bottom.
Our full method fails when background looks like an object and is spatio-temporally more consistent than the object (a, c), or the boundaries of motion clusters include the multiple objects or background together (b, d--e).
The localization results in (b) and (f) are reasonable although they are incorrect according to the P{\small ASCAL} criterion.
(Best viewed in color.)}
\label{fig:qualitative_coloc_bad}
\end{figure*}

Sample qualitative results are shown in Fig.~\ref{fig:qualitative_coloc}~and~\ref{fig:qualitative_coloc_bad}, where the regions localized by our full model are compared with those of F(A), which relies only on image-based information.
F(A) already outperforms the previous state of the art, but its results are often temporally inconsistent when the object undergoes severe pose variation or multiple target objects exist in a video. 
We handle this problem by enforcing temporal consistency on the solution.


\subsection{Unsupervised object discovery and tracking}

In the unsupervised setting, where videos with different object classes are all mixed together, our method still outperforms existing video colocalization techniques even though it does not use any supervisory information, as summarized in Table~\ref{tab:corloc}. 
It performs slightly worse than the state of the art in video segmentation~\cite{Papazoglou2013}, which uses a foreground/background appearance model.  
Note however that (1) such a video-specific appearance model would probably further improve our localization accuracy; 
and (2) our method attacks a more difficult problem, and, unlike \cite{Papazoglou2013}, discovers the underlying topology of the video collection.

\begin{figure}[!htbp]
\centering
    \includegraphics[width = 0.99 \linewidth]{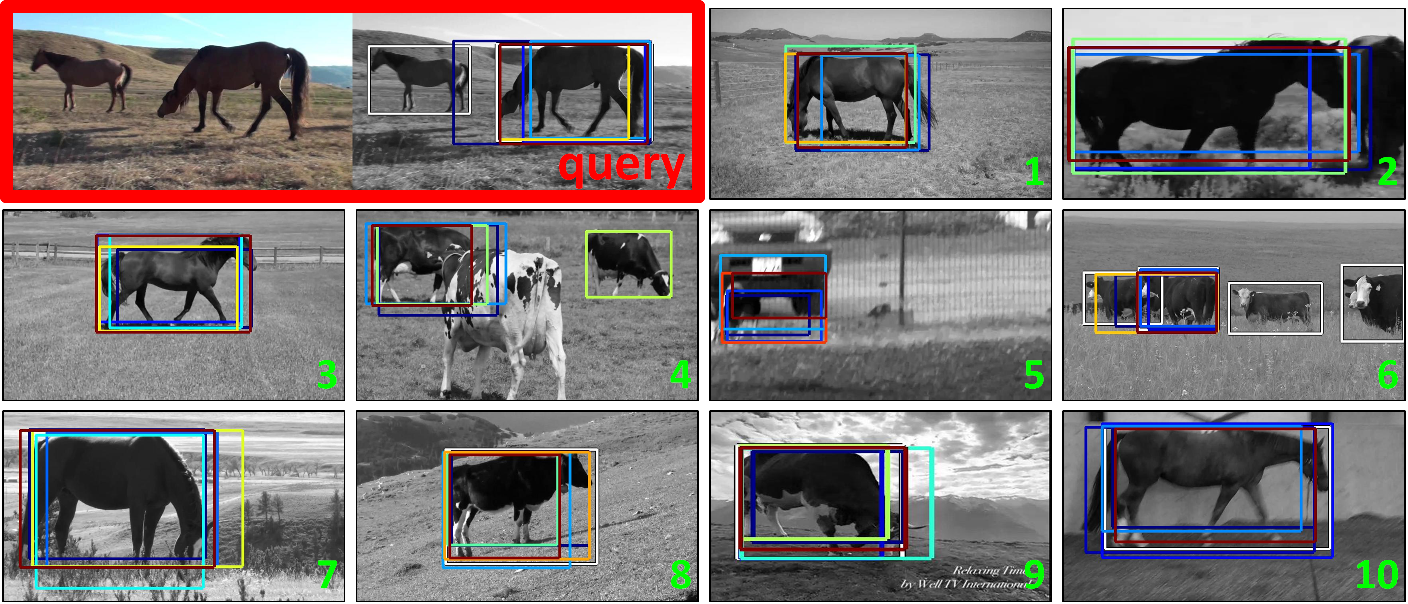}
\caption{A query frame (bold outer box) from the ``horse'' class and its nearest neighbor frames at the last iteration of the unsupervised object discovery and tracking.
The appearances of top-5 object candidates (inner boxes) of the nearest neighbors look similar with those of the query, 
although half of them come from the ``cow'' class (4th, 6th, 8th, and 9th) or the ``car'' class (5th).
}
\label{fig:retr_NN}
\end{figure}

\begin{figure}[!htbp]
\centering
    \includegraphics[width = 0.9 \linewidth]{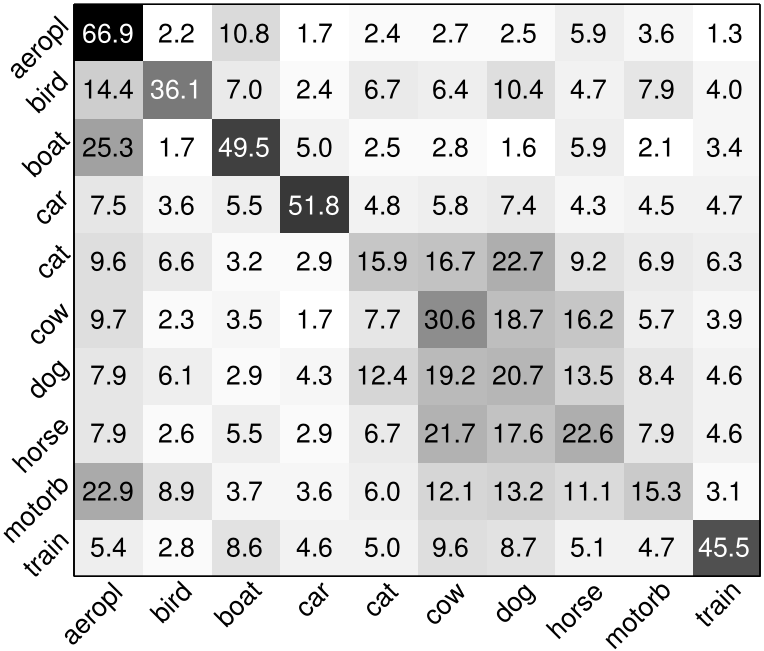}
\caption{Confusion matrix of nearest neighbor retrieval.
Rows correspond to query classes and columns indicate retrieved classes.
Diagonal elements correspond to the CorRet values on Table~\ref{tab:corret}.
}
\label{fig:confusion_mat}
\end{figure}

The quality of nearest-neighbor retrieval is measured by CorRet and quantified in Table~\ref{tab:corret}.
Even in the case where some neighbors do not come from the same class as the query, object candidates in the neighbor frames usually resemble to those in the query frame, as illustrated in Fig.~\ref{fig:retr_NN}.
To illustrate the recovered topology between classes, we provide a confusion matrix of the retrieval results in Fig.~\ref{fig:confusion_mat}, 
showing that most classes are most strongly connected to themselves, and some classes with similar appearances (\eg, ``cat'', ``dog'', ``cow'', and ``horse'') have to some extent connections between them.
Finally, we measure the accuracy of nearest neighbor classification that is based on neighbor frames provided by our method and their ground-truth labels.
The classification accuracy in top-1 and top-2 error rates is summarized in Table~\ref{tab:corret}. 
The error rates are low when the query class usually shows unique appearances (\eg, ``aeroplane'', ``boat'', ``car'', and ``train''), while high if there are other classes with similar appearances (\eg, ``cat'', ``dog'', ``cow'', and ``horse'').


\begin{table*}[!htbp]
\begin{center}
\caption{CorRet scores and top-$k$ error rates of our method on the YouTube-Object dataset in the fully unsupervised setting.}
\label{tab:corret}
\begin{tabular}{c|cccccccccc|c}
\Xhline{3\arrayrulewidth}
	Metric & aeroplane & bird & boat & car & cat & cow & dog & horse & motorbike & train & Avg. \\
\hline
\hline
	CorRet					& 66.9 & 36.1 & 49.5 & 51.8 & 15.9 & 30.6 & 20.7 & 22.6 & 15.3 & 45.5 & 35.5 \\
\hline
	Top-1 error rate		& 12.1 & 51.9 & 34.1 & 25.0 & 84.2 & 45.7 & 70.2 & 73.4 & 83.0 & 33.6 & 51.3 \\
	Top-2 error rate		& \ \ 4.6 & 46.2 & 10.9 & 18.8 & 60.9 & 24.4 & 41.1 & 49.2 & 63.0 & 20.7 & 34.0 \\
\Xhline{3\arrayrulewidth}
\end{tabular}
\end{center}
\label{tab:corloc_voc}
\vspace{-0.6cm}
\end{table*}

\section{Discussion and Conclusion}
\label{sec:conclusion}

We have proposed a novel approach to localizing objects 
in an unlabeled video collection by a combination of object discovery and tracking. 
Not only does our method find objects in each video, 
it also reveals a network structure associating frames and objects across videos. 
It alternatively optimizes the localization objective and the neighborhood structure, improving each. 
We have demonstrated the effectiveness of the proposed method on the YouTube-Object dataset, 
where it significantly outperforms the state of the art in colocalization even though it uses much less supervision.
Some issues still remain for further exploration. 
As it stands, our method is not appropriate for videos 
with a single dominant background and highly non-rigid object (\eg, the UCF-sports dataset).
Next on our agenda is to address these issues, 
using for example video stabilization and foreground/background models~\cite{Papazoglou2013,Lee2011key}.


\vspace{-0.3cm}
\paragraph{Acknowledgments.}
This work was supported by the ERC grants Activia, Allegro, and VideoWorld, and the Institut Universitaire de France. 

{\small
\bibliographystyle{ieee}
\bibliography{vod}
}

\end{document}